\documentclass[conference]{IEEEtran}
\IEEEoverridecommandlockouts
\usepackage{cite}
\usepackage{amsmath,amssymb,amsfonts}
\usepackage{algorithm}
\usepackage{algorithmic}
\usepackage{graphicx}
\usepackage[caption=false,font=footnotesize]{subfig}
\usepackage{textcomp}
\usepackage{xcolor}
\usepackage{soul}

\def\BibTeX{{\rm B\kern-.05em{\sc i\kern-.025em b}\kern-.08em
    T\kern-.1667em\lower.7ex\hbox{E}\kern-.125emX}}

\begin{document}

\title{Texture-Aware Strategies for Subregion Style Transfer Augmentation}

\author{
\IEEEauthorblockN{
Cheng Ji\textsuperscript{1},
Yu Zhao\textsuperscript{1},
Xiaochong Jiang\textsuperscript{2},
Wenting Yang\textsuperscript{3},
Shiqi Yang\textsuperscript{4}
}
\IEEEauthorblockA{
\textsuperscript{1}\textit{University of Illinois Urbana-Champaign}, Champaign, USA\\
\textsuperscript{2}\textit{Northeastern University}, Boston, USA\\
\textsuperscript{3}\textit{University of California San Diego}, San Diego, USA\\
\textsuperscript{4}\textit{New York University}, New York, USA\\
chengji5@illinois.edu,
yuzhao5@illinois.edu,
jiang.xiaoc@northeastern.edu,
wey023@ucsd.edu,
sy3506@nyu.edu
}
}
\maketitle

\begin{abstract}
A common strategy for data augmentation is to apply style transfer on training images to enrich appearance variations. Recent approaches have combined style augmentation with random erasing by selectively replacing image subregions with style-transferred patches. However,  existing methods rely on uniformly random region selection or fixed geometric constraints, which may fail to cover informative texture patterns and limit augmentation diversity. To address these limitations, we propose a texture-aware subregion style augmentation framework that prioritizes regions with strong texture responses for style replacement. Furthermore, we extend this framework with a hybrid region strategy that combines a texture-driven region with an auxiliary random region, implicitly enabling more diverse style perturbations. Experiments demonstrate that our approach achieves higher classification accuracy and improves robustness compared with existing style-based augmentation methods.

\end{abstract}


\section{Introduction}

Deep neural networks have demonstrated impressive performance on a wide range of computer vision tasks \cite{lecun1995convolutional}, including visual question answering and multimodal deception detection \cite{li2024exploring, li2024deception}. However, their success largely relies on the availability of large-scale and diverse labeled datasets \cite{coates2011analysis, hastie2009unsupervised, kumar2024image}. In many practical scenarios, collecting sufficient training data is expensive, time-consuming, or even infeasible, which often results in overfitting and limited generalization ability. Beyond data scarcity, the distribution and quality of training data can also substantially affect learning performance, as demonstrated by recent large-scale benchmarks on class-imbalanced learning \cite{liu2025climb}.

Data augmentation is a widely adopted technique to alleviate this problem by artificially increasing the diversity of training data through generative
modeling or label-preserving transformations \cite{ding2022data, kumar2024image2, maharana2022review}. Generative methods refer to artificial construction of new data, with usage of generative models like Generative Adversarial Networks (GANs) and Large Language Models (LLMs) \cite{wang2025selfdestructivelanguagemodel, wang2025reasoningretrievalstudyanswer, ge2025electronic, liu2025research, guo2025classification, guo2025metabolitechat}. Label-preserving transformations refer to performing transformations on existing samples. It exposes models to a broader set of variations and encourages them to learn more robust representations. Common strategies include flipping, cropping, rotation and brightness adjustment \cite{zhong2020random, lu2023machine, yang2024comparative}.

Nevertheless, these traditional methods frequently fail to fully capture the visual complexity of real-world environments, driving active research into more advanced frameworks. Among them, style-based augmentation has emerged as an effective way to improve robustness against appearance variations \cite{atapour2018real, jackson2019style, chen2016fast, xu2024style}. By altering the artistic style of an image while preserving its semantic content, style augmentation enables models to become less sensitive to superficial appearance changes and more focused on structural cues. Recent studies on vision-language models further demonstrate that even semantics-preserving visual perturbations can substantially alter model behavior, highlighting the broader importance of robustness to superficial visual variations \cite{liu2026vlms}.

Most existing style augmentation methods operate at the image level, applying stylization to the entire image \cite{jackson2019style}. However, global style transformation may not always be optimal. Strong stylization over the whole image can distort important structures and introduce unnecessary artifacts, which may degrade discriminative features. To address this issue, recent work has proposed applying style transfer only to a subregion of the image, thereby preserving global semantics while injecting localized appearance variations \cite{yang2025data}.

Despite its effectiveness, existing subregion style transfer methods still suffer from two major limitations \cite{yang2025data}. First, the location of the subregion is selected uniformly at random, without considering the underlying image content. Consequently, stylization may be applied to uninformative regions, resulting in limited benefit. Second, the subregion shape is fixed, which constrains the diversity of spatial perturbations and reduces the expressiveness of the augmentation.

Motivated by this insight, we propose a subregion style transfer augmentation that addresses the inefficiencies of random selection while enhancing shape diversity in a simple yet effective mechanism. Specifically, our strategy determines \emph{where} to apply style perturbations in a content-aware manner. We observe that stylizing texture-rich regions is more likely to introduce meaningful appearance variations, while random selection often wastes augmentation budget on homogeneous areas \cite{geirhos2018imagenet}. At the same time, overly deterministic selection may reduce spatial diversity. These considerations motivate a selection mechanism that balances informativeness and diversity. 

\begin{figure}[htbp]
  \centering
  \subfloat[]{\includegraphics[width=0.22\linewidth]{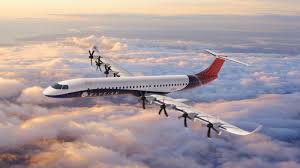}}
  \hfil
  \subfloat[]{\includegraphics[width=0.22\linewidth]{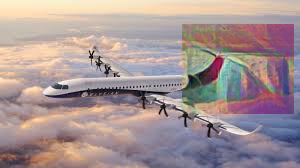}}
  \hfil
  \subfloat[]{\includegraphics[width=0.22\linewidth]{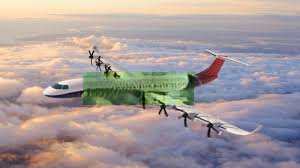}}
  \hfil
  \subfloat[]{\includegraphics[width=0.22\linewidth]{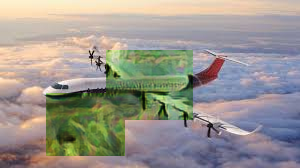}}
  
  \subfloat[]{\includegraphics[width=0.22\linewidth]{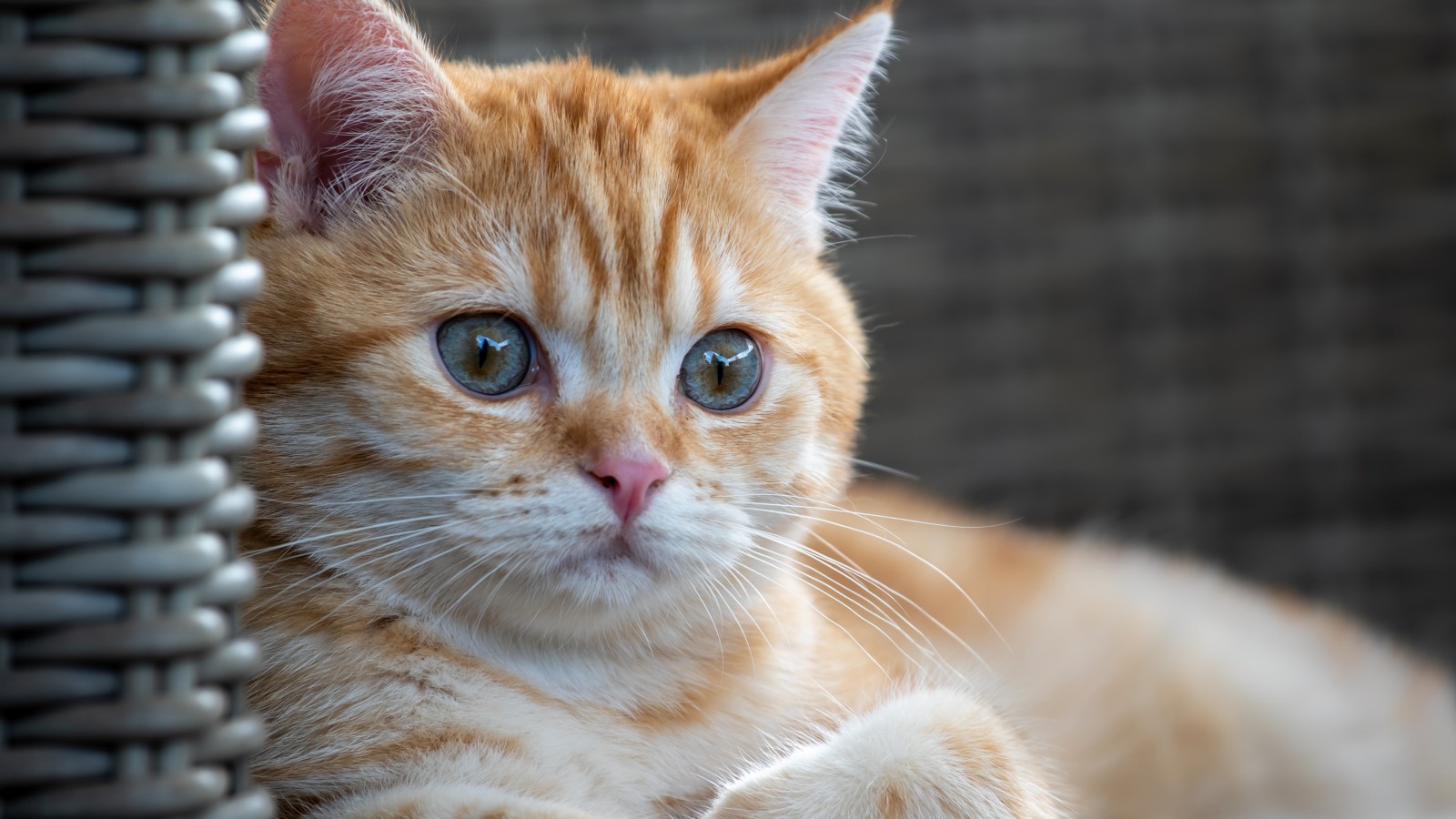}}
  \hfil
  \subfloat[]{\includegraphics[width=0.22\linewidth]{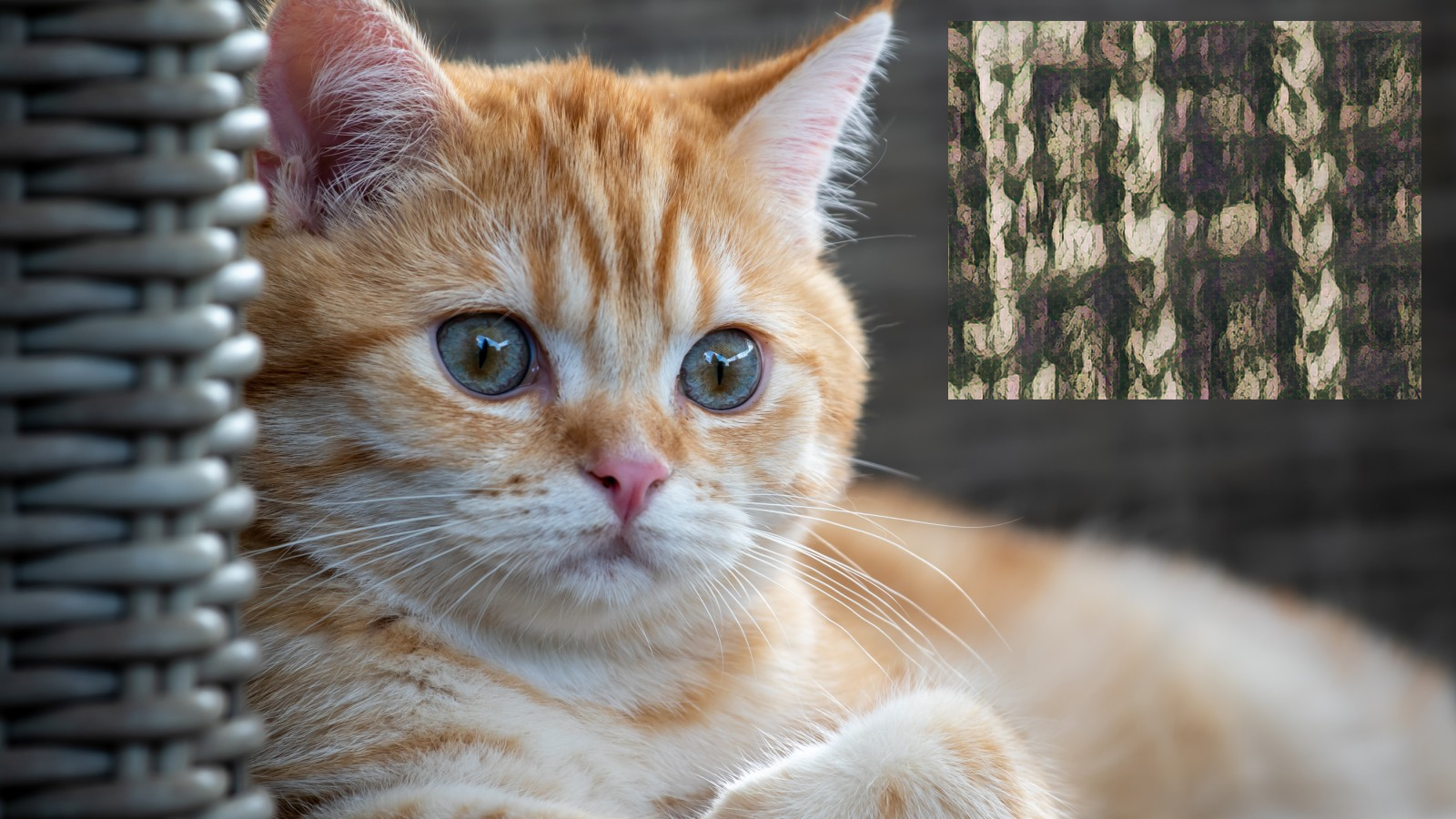}}
  \hfil
  \subfloat[]{\includegraphics[width=0.22\linewidth]{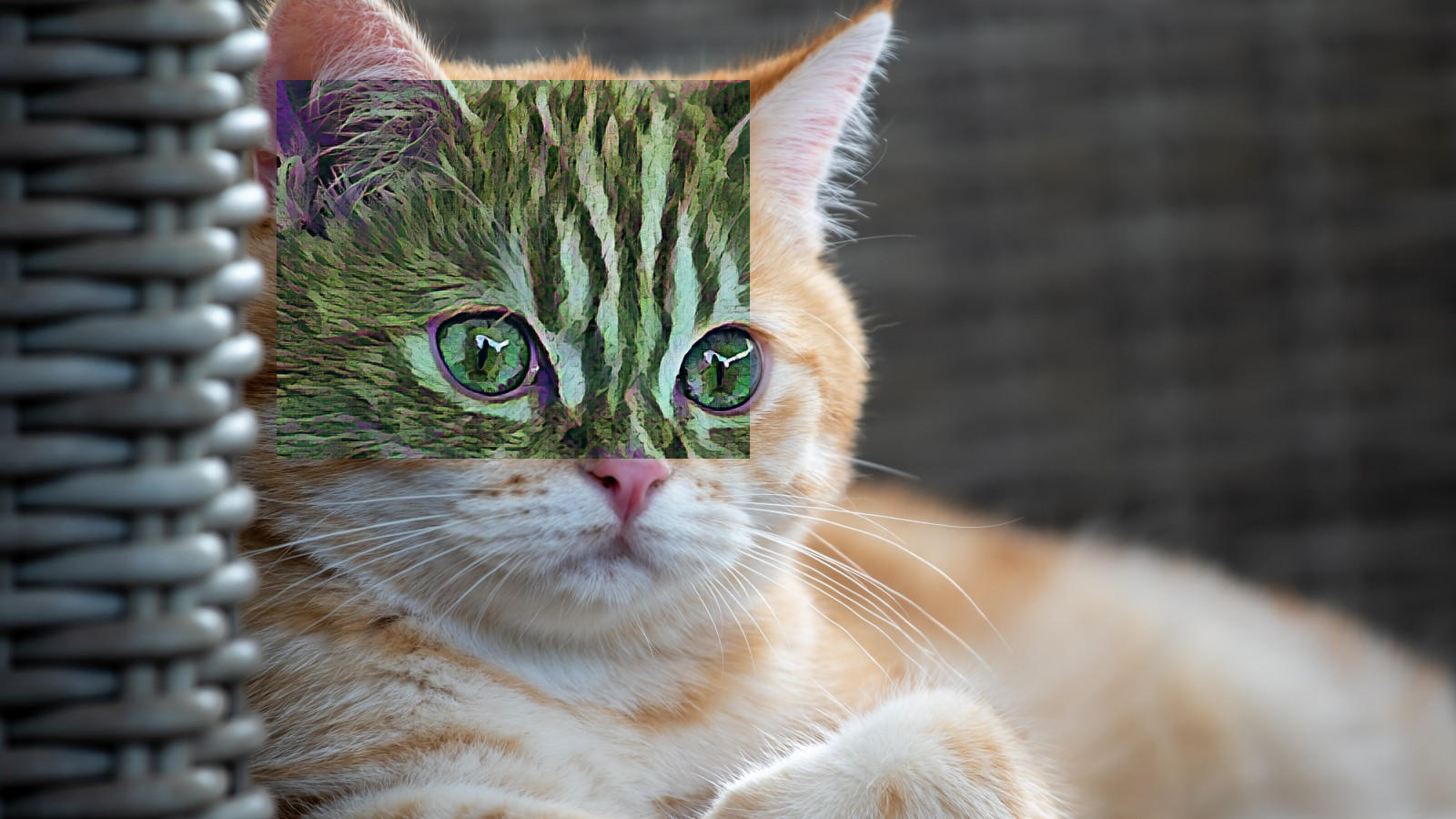}}
  \hfil
  \subfloat[]{\includegraphics[width=0.22\linewidth]{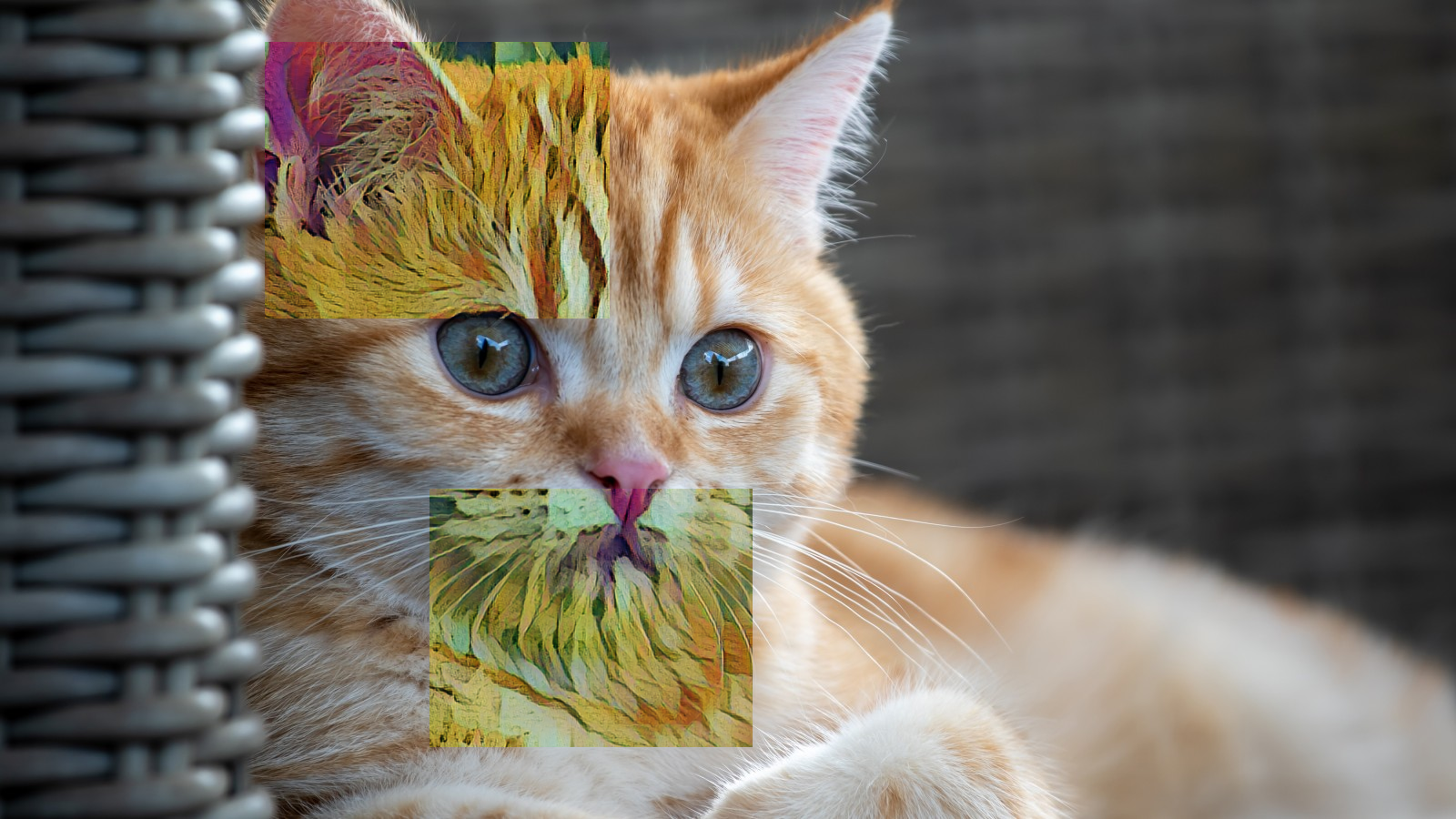}}
  
  \caption{Examples of different subregion style transfers: (a)(e) Input Image (b)(f) Random subregion method (c)(g) Texture-intensive subregion method (d)(h) Mixed subregion method.}
  \label{fig:cmp-study}
\end{figure}

The main contributions of our work are summarized as follows:
\begin{itemize}
  \item  We propose two subregion style transfer frameworks for style-based augmentation: (i) a texture-aware method that prioritizes regions with strong texture responses, and (ii) a hybrid strategy that combines a texture-driven region with an randomly sampled region to increase the diversity of perturbations. Both strategies can be integrated with any style transfer technique, as illustrated in Figure \ref{fig:cmp-study}.
  \item We identify and explicitly address the limitations of existing random style transfer methods. By deliberately targeting dominant local texture cues, our approach effectively disrupts the model's tendency to overfit.
  \item  We conduct comprehensive experiments on STL-10 and show that our approach consistently improves classification accuracy and robustness across different network capacities compared with existing style-based augmentation baselines.
\end{itemize}

\section{Methods}
In this section, we present our texture-aware subregion style transfer strategies in detail. We first describe the overall subregion style transfer process and then introduce our subregion selection strategies.

\subsection{Subregion Style Transfer}
Subregion style transfer consists of two sequential processes: subregion selection and region-level style transformation. Given an input image, a target subregion is first determined according to a specified selection strategy. Subsequently, a style transfer operation is applied to the selected region, while the remaining areas of the image are kept unchanged. The overall procedure is described in  Alg. \ref{algorithm 1}.

Style transfer refers to a class of image processing techniques that modify the visual appearance of an image while preserving its underlying semantic structure. To be suitable for data augmentation, the adopted style transfer method must be both computationally efficient and capable of producing diverse and stochastic style variations \cite{jackson2019style}. Following this principle, we employ the approach of Jackson et al., which injects randomness on the fly.

\begin{algorithm}[htbp]
\caption{Subregion Style Transfer}
\label{algorithm 1}
\begin{algorithmic}[1] % [1] 表示显示行号，如果不需要可以去掉
\renewcommand{\algorithmicrequire}{\textbf{Input:}}
\renewcommand{\algorithmicensure}{\textbf{Output:}}
\REQUIRE Input image $I$; Augmentation probability $p$; Subregion selection method $f$
\ENSURE Augmented image $I^{\ast}$.
\STATE \textbf{Initialization:} $p_1 \leftarrow $ Rand(0, 1), $I^{\ast} \leftarrow I$;
\IF{$p_1 \geq p$}
    \RETURN $I^{\ast}$
\ELSE
    \STATE $I' \leftarrow \text{StyleTransfer}(I)$;
    \STATE $S \leftarrow f(I)$;
    \FOR{each $s \in S$}
        \STATE Replace $I^{\ast}[s] \leftarrow I'[s]$
    \ENDFOR
    \RETURN $I^{\ast}$
\ENDIF
\end{algorithmic}
\end{algorithm}

\label{subsec:texture_aware_selection}
To determine an informative region for style transfer, we propose a \emph{texture-aware subregion selection} strategy. The core intuition is that image regions with rich local structures (e.g., edges, corners, and fine details) are often more informative and can benefit more from style transformation than textureless or flat regions \cite{geirhos2018imagenet}. Therefore, we aim to place the stylization patch on areas exhibiting high texture intensity. The overall procedure is described in  Alg. \ref{alg:texture_aware_patch}.

Given an input image
$I \in \mathbb{R}^{C \times H \times W}$, we first convert it into a grayscale
image $G \in \mathbb{R}^{1 \times H \times W}$. A texture map
$T \in \mathbb{R}^{H \times W}$ is then computed using the magnitude of the
Laplacian response:

\begin{equation}
T = \left| \Delta(G) \right|,
\end{equation}

where $\Delta(\cdot)$ denotes the discrete Laplacian operator and
$|\cdot|$ indicates the element-wise absolute value. Pixels with larger values
in $T$ correspond to stronger local intensity variations.

Exhaustively evaluating all possible patch locations is computationally
expensive. Instead, we adopt a stochastic search strategy. Specifically, we
uniformly sample $N$ valid candidate top-left coordinates
$\{(y_i, x_i)\}_{i=1}^{N}$. For each candidate, we compute the mean texture score inside the
patch region:

\begin{equation}
s_i = \frac{1}{hw}
\sum_{u = y_i}^{y_i + h - 1}
\sum_{v = x_i}^{x_i + w - 1}
T(u,v).
\end{equation}

The candidate with the maximum score is selected as the final subregion:

\begin{equation}
i^{\ast} = \arg\max_{i} \; s_i.
\end{equation}

The corresponding patch bounding box
$(y_{i^{\ast}}, x_{i^{\ast}}, y_{i^{\ast}} + h, x_{i^{\ast}} + w)$ is then used
for subsequent style transfer.

\begin{algorithm}[htbp]
\caption{Texture-aware Subregion Selection}
\label{alg:texture_aware_patch}
\begin{algorithmic}[1]
\renewcommand{\algorithmicrequire}{\textbf{Input:}}
\renewcommand{\algorithmicensure}{\textbf{Output:}}
\REQUIRE Input image $I$; patch size $(h,w)$; number $N$
\ENSURE Patch region $S$
\STATE $G \leftarrow \textsc{ToGray}(I)$
\STATE $T \leftarrow \left| \Delta(G) \right|$
\STATE Sample $\{(y_i, x_i)\}_{i=1}^{N}$ uniformly from $[0, H-h] \times [0, W-w]$
\FOR{$i = 1$ \textbf{to} $N$}
  \STATE $R_i \leftarrow T[y_i : y_i + h, \; x_i : x_i + w]$
  \STATE $s_i \leftarrow \textsc{Mean}(R_i)$
\ENDFOR
\STATE $i^{\ast} \leftarrow \arg\max_i s_i$
\RETURN $S = (y_{i^{\ast}}, x_{i^{\ast}}, y_{i^{\ast}} + h, x_{i^{\ast}} + w)$
\end{algorithmic}
\end{algorithm}

\subsection{Hybrid Subregion Selection}
\label{subsec:hybrid_selection}

While texture-rich selection places stylization on highly informative regions, relying solely on such deterministic criterion may reduce spatial
diversity and lead to biased augmentation. To balance \emph{exploration} and \emph{exploitation}, we propose a \emph{hybrid subregion selection} strategy that combines random sampling with texture-aware selection.

Given an input image
$I \in \mathbb{R}^{C \times H \times W}$ and patch sizes $(h_1,w_1),(h_2,w_2)$, we select two
candidate subregions. The first region is obtained by uniform random sampling,
denoted as $R^{\text{rand}}$, which encourages spatial diversity. The second
region is obtained by the texture-aware method described in
Section~\ref{subsec:texture_aware_selection}, denoted as
$R^{\text{tex}}$, which prefers regions with strong local structures.

The final hybrid region set is defined as

\begin{equation}
\mathcal{R} = \{R^{\text{rand}}, R^{\text{tex}}\}
\end{equation}

Both regions in $\mathcal{R}$ are subsequently used for style transfer, which
introduces both spatial diversity and structure-aware perturbations.  The overall procedure is described in  Alg. \ref{alg:hybrid_selection}.

\begin{algorithm}[htbp]
\caption{Hybrid Subregion Selection}
\label{alg:hybrid_selection}
\begin{algorithmic}[1]
\renewcommand{\algorithmicrequire}{\textbf{Input:}}
\renewcommand{\algorithmicensure}{\textbf{Output:}}
\REQUIRE Input image $I$; patch size $(h_1,w_1), (h_2, w_2)$
\ENSURE Patch Region $S$
\STATE Sample $(y_r, x_r)$ uniformly from $[0, H-h_1] \times [0, W-w_1]$
\STATE $R^{\text{rand}} \leftarrow (y_r, x_r, y_r + h_1, x_r + w_1)$
\STATE $R^{\text{tex}} \leftarrow f_{\text{tex}}(I, h_2, w_2)$ \COMMENT{Algorithm~\ref{alg:texture_aware_patch}}
\RETURN $\{R^{\text{rand}}, R^{\text{tex}}\}$
\end{algorithmic}
\end{algorithm}

\section{Experiment}
\subsection{Dataset}
We evaluated our proposed method on the STL-10 dataset. This dataset comprises 5,000 labeled training images and 8,000 test images across 10 classes, with a resolution of $96 \times 96$ pixels \cite{coates2011analysis}.

The choice of the STL-10 dataset is motivated by two key factors. First, its limited number of labeled training samples poses a high risk of overfitting when training models like ResNet-50, making it a rigorous benchmark for evaluating overfitting behavior. Second, the $96 \times 96$ image resolution preserves sufficient high-frequency details, which is essential for analyzing texture dependence and distinguishes STL-10 from lower-resolution datasets such as CIFAR. 

\subsection{Experiment Settings}\label{AA}
We consider supervised \emph{image classification} as the primary evaluation task. To comprehensively evaluate the scalability of our method, we employ the ResNet family as our backbone architectures, specifically ResNet-18, ResNet-50, and ResNet-101. This selection allows us to analyze the impact of texture-guided regularization across varying model capacities, ranging from shallow networks to deep, over-parameterized architectures. All models are trained from scratch without pre-training to isolate the specific contributions of our augmentation strategy.

We benchmark our approach against the random subregion style augmentation. To ensure a rigorous comparison, the baseline is configured to match the total expected occlusion area of our method ($E[\text{occlusion area ratio}] \approx 0.4$). This control ensures that any performance gain is attributable to our texture-guided spatial strategy rather than the amount of information transformed.

In all experiments, we utilize the Adam optimizer with momentum $\beta_1$ = 0.5, $\beta_2$ = 0.999, and an initial learning rate of 0.001, aligning with the settings of other data augmentation methods \cite{jackson2019style, zhong2020random, kingma2014adam}. The style interpolation parameter is set to $\alpha = 0.8$ and augmentation ratio is set as 1:1 \cite{jackson2019style}. All experiments are trained on an NVIDIA RTX 5090 GPU for 100 epochs. 

The experimental groups are as follows:
\begin{enumerate}
    \item Naive augmentation: the original dataset with standard geometric augmentations (e.g., random crop, horizontal flip). To maintain consistency in epoch size/iterations with the augmented groups, this dataset is oversampled (copied and stacked) to match the 1:1 ratio
    \item Random subregion style transfer: subregions are replaced with style-transferred patches at random locations
    \item Texture-intensive subregion style transfer: texture-aware method described in Section~\ref{subsec:texture_aware_selection}
    \item Mixed subregion style transfer: the hybrid approach described in Section~\ref{subsec:hybrid_selection}
\end{enumerate}

\subsection{Results}
\begin{figure}[htbp]
  \centering
    \includegraphics[width=0.9\linewidth, height=0.6\linewidth]{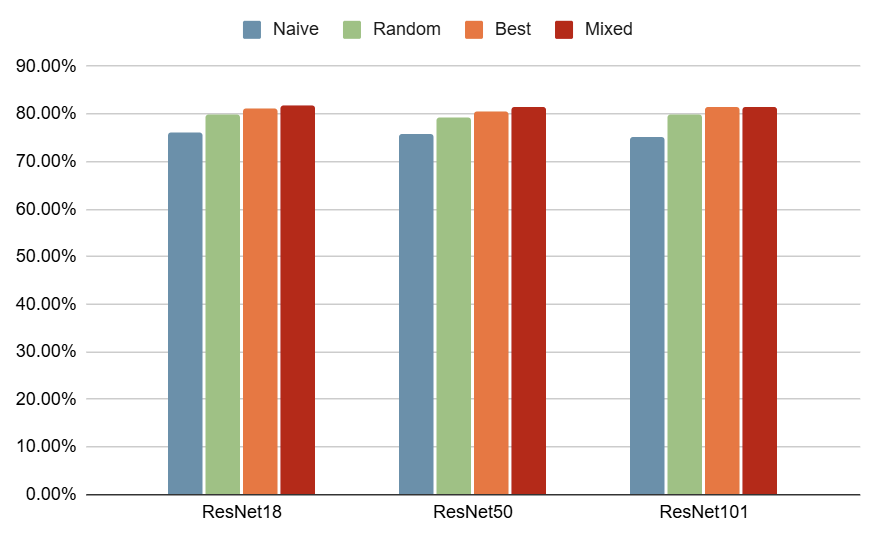}
    \caption{Classification Accuracy of ResNets on STL-10 test set. ``Naive'' represents naive augmentation. ``Random" represents random subregion style transfer.
    ``Best" represents texture-intensive subregion style transfer. ``Mixed" represents mixed subregion style transfer.}
  \label{fig:bar}
\end{figure}

As shown in Fig. \ref{fig:bar}, our mixed method achieves a top-1 accuracy of $81.7\%$ on ResNet-50 in just 100 epochs. This result is significantly more efficient than other methods. Additionally, it is better than the $80.8\%$ baseline achieved using $100,000$ epochs reported by Jackson et al. \cite{jackson2019style}. This performance advantage is consistent across ResNet-18, ResNet-50, and ResNet-101, demonstrating the method's robustness across varying model capacities.

\begin{figure}[htbp]
  \centering
    \includegraphics[width=0.8\linewidth, height=0.6\linewidth]{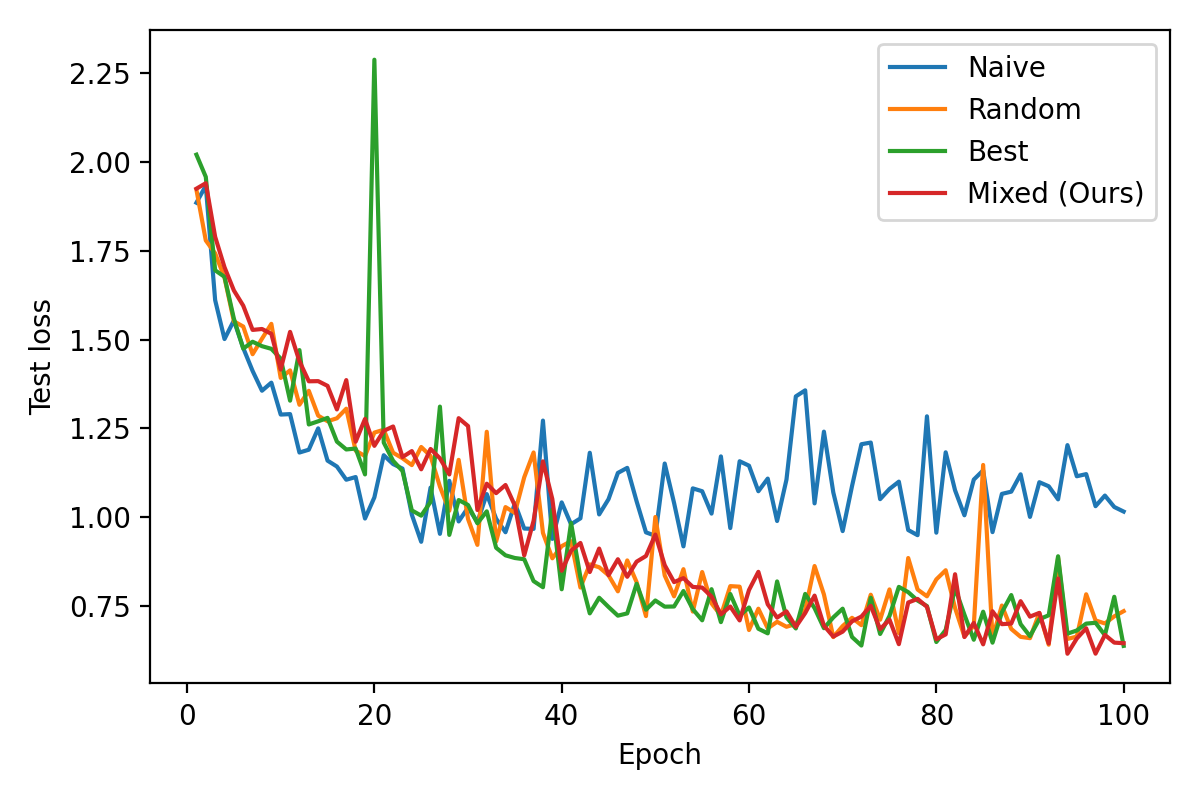}
    \caption{Test loss of ResNet101 on STL-10 test set.}
  \label{fig:loss}
\end{figure}

We further analyzed the test loss trajectories to assess training stability (Fig.~\ref{fig:loss}). The \textit{Naive} baseline exhibits larger oscillations and converges to a visibly higher loss level in the later stage. The \textit{Random} and \textit{Best} strategies generally achieve lower test loss, but more occasional abrupt spikes are observed (e.g., a pronounced spike for \textit{Random} around
epoch 80), suggesting higher epoch-to-epoch variance. In contrast, our
\textit{Mixed} strategy maintains a smoother loss descent and stays
consistently low after convergence, indicating more reliable training behavior.

Another key benefit of our approach is reflected in the train--test accuracy gap, especially for over-parameterized models such as ResNet-50 trained on STL-10. As shown in Table~\ref{tab:train_test_gap}, our mixed strategy reduces the gap to $<1.1\%$, compared to the nearly $3\%$ gap produced by  random style transfer.

This tight convergence indicates that the model is not memorizing training samples. Instead, the difficulty imposed by our style transfer method drives the model to learn more robust structural representations that generalize much better to unseen data.

The success of our approach comes from its unique texture-aware mechanism. Unlike blind random selection, our texture-driven strategy precisely targets and disrupts dominant local features, compelling the model to abandon superficial shortcuts. Furthermore, the hybrid strategy effectively balances deterministic targeting with stochastic exploration. Consequently, our method achieves a remarkably smooth training trajectory and prevents severe overfitting.

\begin{table}[htbp]
\caption{Train--Test Accuracy Gap on STL-10 (ResNet-50).}
\begin{center}
\begin{tabular}{|l|c|c|c|}
\hline
\textbf{Augmentation} & \textbf{Train Acc.} & \textbf{Test Acc.} & \textbf{Gap} $\downarrow$ \\
\hline
Naive & 0.9064 & 0.7605 & 0.1459 \\
\hline
Random Style Transfer & 0.8195 & 0.7921 & 0.0274 \\
\hline
Mixed Strategy (ours) & 0.8275 & \textbf{0.8170} & \textbf{0.0105} \\
\hline
\end{tabular}
\label{tab:train_test_gap}
\end{center}
\end{table}

\section{Conclusions}
In this paper, we presented two texture-guided strategies to address the challenge of training over-parameterized models on limited data. By systematically targeting high-texture regions while maintaining stochastic diversity, our approach successfully disrupts the model's reliance on superficial shortcuts and drives it to learn more robust structural representations. Experiments on STL-10 dataset demonstrate that our method significantly outperforms random subregion transfer baselines and scales effectively across multiple ResNet architectures.

\section{Future Work}
Future work should further examine the generality of texture-rich subregion style transfer by evaluating it on additional datasets and a wider range of vision tasks, in order to characterize when it is most effective and where it may have limitations. Beyond the current texture-aware and hybrid designs, more diverse subregion selection mechanisms also require investigation to improve both the informativeness and diversity of the perturbed regions. Finally, since style transfer is applied only to localized regions, systematic benchmarks on efficiency (e.g., augmentation latency and end-to-end training throughput) are needed to quantify the practical speed advantages over full-image stylization.

\bibliographystyle{IEEEtran}
\bibliography{main}

\end{document}